\documentclass[a4paper,fleqn]{cas-dc}

\usepackage[authoryear,longnamesfirst]{natbib}

\usepackage{algorithm}
\usepackage{algorithmicx}
\usepackage{algpseudocode}
\usepackage{amsmath}

\usepackage{indentfirst}

\floatname{algorithm}{}

\usepackage{array}
\usepackage[utf8]{inputenc}
\usepackage[T1]{fontenc}

\hypersetup{hypertex=true,
colorlinks=true,
linkcolor=blue,
anchorcolor=blue,
citecolor=blue}

\def\tsc#1{\csdef{#1}{\textsc{\lowercase{#1}}\xspace}}
\tsc{WGM}
\tsc{QE}
\tsc{EP}
\tsc{PMS}
\tsc{BEC}
\tsc{DE}
\begin{document}
\let\WriteBookmarks\relax
\def\floatpagepagefraction{1}
\def\textpagefraction{.001}
\let\printorcid\relax

\shorttitle{}

\shortauthors{}

\title [mode = title]{Simple Network Graph Comparative Learning}

\author[1]{Qiang Yu}
% \author[1,3]{CV Radhakrishnan}[type=editor,auid=000,bioid=1,prefix=Sir,role=Researcher,orcid=0000-0001-7511-2910]
\credit{Conceptualization of this study, Methodology, Software.}

\author[1]{Xinran Cheng}
% \author[1,3]{CV Radhakrishnan}[type=editor,auid=000,bioid=1,prefix=Sir,role=Researcher,orcid=0000-0001-7511-2910]
\credit{Methodology, Supervision, Writing-review \& editing.}

\author[2]{Shiqiang Xu}
%[style=chinese]
\credit{Methodology, Supervision, Writing-review \& editing.}
% \address{HIT,Harbin 100000, China}

\author[1]{Chuanyi Liu}
%[style=chinese]
\credit{Methodology, Supervision, Writing-review \& editing.}

\address[1]{organization={School of Computer Science and Technology, Harbin Institute of Technology,Shenzhen},
    city={Shenzhen},
    % citysep={}, % Uncomment if no comma needed between city and postcode
    postcode={518071},
    country={China}}

\address[2]{organization={Inspur Cloud Information Technology Co., Ltd},
    city={Jinan},
    % citysep={}, % Uncomment if no comma needed between city and postcode
    postcode={250101},
    country={China}}

% % Address/affiliation
% \affiliation[2]{organization={Sayahna Foundation},
%     % addressline={}, 
%     city={Jagathy},
%     % citysep={}, % Uncomment if no comma needed between city and postcode
%     postcode={695014}, 
%     state={Trivandrum},
%     country={India}}

% \author%
% [1,3]
% {Rishi T.}
% \cormark[2]
% \fnmark[1,3]
% \ead{rishi@stmdocs.in}
% \ead[URL]{www.stmdocs.in}
% \affiliation[3]{organization={STM Document Engineering Pvt Ltd.},
%     addressline={Mepukada}, 
%     city={Malayinkil},
%     % citysep={}, % Uncomment if no comma needed between city and postcode
%     postcode={695571}, 
%     state={Trivandrum},
%     country={India}}

\cortext[cor1]{Corresponding author}
% \cortext[cor2]{Principal corresponding author}

% \fntext[fn1]{This is the first author footnote. but is common to third
% \fntext[fn2]{Another author footnote, this is a very long footnote and
%   it should be a really long footnote. But this footnote is not yet
%   sufficiently long enough to make two lines of footnote text.}

% % For a title note without a number/mark
% \nonumnote{This note has no numbers. In this work we demonstrate $a_b$
%   the formation Y\_1 of a new type of polariton on the interface
%   between a cuprous oxide slab and a polystyrene micro-sphere placed

\begin{abstract}
The effectiveness of contrastive learning methods has been widely recognized in the field of graph learning, especially in contexts where graph data often lack labels or are difficult to label. However, the application of these methods to node classification tasks still faces a number of challenges. First, existing data enhancement techniques may lead to significant differences from the original view when generating new views, which may weaken the relevance of the view and affect the efficiency of model training. Second, the vast majority of existing graph comparison learning algorithms rely on the use of a large number of negative samples. To address the above challenges, this study proposes a novel node classification contrast learning method called Simple Network Graph Comparative Learning (SNGCL). Specifically, SNGCL employs a superimposed multilayer Laplace smoothing filter as a step in processing the data to obtain global and local feature smoothing matrices, respectively, which are thus passed into the target and online networks of the siamese network, and finally employs an improved triple recombination loss function to bring the intra-class distance closer and the inter-class distance farther. We have compared SNGCL with state-of-the-art models in node classification tasks, and the experimental results show that SNGCL is strongly competitive in most tasks.
\end{abstract}

% \includegraphics{figs/grabs.pdf}

% \begin{highlights}
% \item Research highlights item 1
% \item Research highlights item 2
% \item Research highlights item 3
% \end{highlights}

% Keywords

\maketitle

\section{Introduction}

\par{In recent years, graph representation learning\textsuperscript{\citep{park2019symmetric,hamilton2020graph,chen2024exploring,dai2025graph}} methods have been widely developed. Frequently termed graph embedding learning, the goal of this technique is to transform complex structural and semantic information in graph data into low-dimensional feature vectors. Such a transformation makes data analysis and processing in reduced dimensional spaces feasible, and is applicable to areas such as social network analysis\textsuperscript{\cite{peng2020graph}} and information mining\textsuperscript{\cite{blei2003latent}}, for instance. Graph neural networks\textsuperscript{\cite{defferrard2016convolutional}} have notably excelled in this field, they rely on manually labeled training data, which requires significant time and resources, and lacks in robustness and generalization ability. To this end, contrast learning\textsuperscript{\cite{chen2020simple}}, a self-supervised learning technique capable of learning without labeled data, provides a discriminative-based representation learning framework, and the strategy has recently demonstrated excellent performance in the field of representation learning.}
\par{Historically, unsupervised methodologies, exemplified by DeepWalk\textsuperscript{\cite{perozzi2014deepwalk}}, employ random walks for the unsupervised acquisition of graph representations. It treats a sequence of nodes as similar to sentences in a natural language and utilizes the Word2Vec\textsuperscript{\cite{mikolov2013distributed}} model to learn a distributed representation of nodes. Like DeepWalk, Node2Vec\textsuperscript{\cite{grover2016node2vec}} is an improved stochastic wandering method that controls the wandering strategy based on user-defined parameters to better capture the structure and relationships between nodes. However its over-reliance on graph structural proximity. On the other hand graph neural networks not only take into account the structural relationships between nodes but also incorporate the feature information of the nodes into the learning. This makes graph neural networks a widely used method for processing graph data. However, it still has some drawbacks; it needs to rely on labeled nodes or graphs for supervised learning, performs poorly on sparse or incompletely connected graphs, and also has difficulty in capturing long-distance node relationships. Contrastive learning represents a notable paradigm within the realm of unsupervised learning, has been implemented in various algorithms, e.g., DGI\textsuperscript{\cite{velickovic2019deep}}, inspired by the ideas of DIM\textsuperscript{\cite{hjelm2018learning}} in the graph domain and applied to graph data, proposes a general approach to learn node representations in graph-structured data in an unsupervised manner, focusing on enhancing the mutual information between the augmented representation of the graph and the graph information currently extracted. On this basis MVGRL\textsuperscript{\cite{hassani2020contrastive}} proposes to train graph encoders by maximizing the mutual information between the representations encoded in different structural views of the graph. Other approaches such as GRACE\textsuperscript{\cite{zhu2020deep}} learn node representations through enhancing the uniformity of node representations in both attribute-level and structure-level views using node-level comparison goals to improve the mutual information between input node features and high-level node embeddings.}

\par{Although the above approaches have achieved remarkable success, they are affected to some or all extent by the following limitations. First, existing data augmentation approaches\textsuperscript{\cite{flake2004graph,callaway2000network,aggarwal2014evolutionary}} often lead to significant differences between the new view and the original view, affecting view quality and model training effectiveness\textsuperscript{\cite{dong2024effectiveness,zhou2025data}}. At the same time, there is the problem of time-consuming and computationally resource-intensive processing of large-scale data. Secondly, most existing graph comparison learning methods depend on a substantial quantity of negative samples to ensure stability and accuracy, which however increases the training time and computational cost. Recent studies have even begun to question the necessity of such a large number of negatives in graph comparison learning \textsuperscript{\cite{huang2025does}}. Obviously, it is attractive to seek a training method that is reliable and capable of generating high-quality representations, as recently reinforced by negative-free advances that leverage adaptive augmentation and multi-view correlation \textsuperscript{\cite{zhao2022graph,Wang2024Multiview,wang2025prompt}}.

To address these issues, this paper proposes a Simple comparison node classification (SNGCL) method, which improves the existing methods in terms of data enhancement, network structure and comparison loss function. In the data enhancement stage, we utilize multiple Laplace smoothing filters to generate a data view of local and global features, which maintains the structural properties and effectively reduces the noise interference. The main structure of the network adopts a momentum-guided siamese network architecture, which facilitates the online network to explore better-quality representations, while avoiding representation collapse by avoiding the use of a large number of negative samples, thus gaining advantages in enhancing classification effectiveness and training efficiency. Further, we employ a new contrast objective function that does not rely on the discriminator and exploits the complementarity between them by integrating structural and neighborhood information with the aim of increasing inter-class variability. Meanwhile, an upper-bound loss function is employed to reduce the intra-class variability and additionally improve the model's effectiveness.}

\par{
This paper's key contributions are outlined as follows.
}

\begin{itemize} \item We propose a reliable and capable of generating high-quality representations a simple comparison node classification method called SNGCL.
\item  In this paper, we design a filter consisting of superimposed multilayer normalized and symmetric normalized Laplace filters to generate global and local views. With this filter, the high-frequency noise in the attributes is filtered out and the structural properties are preserved, thus improving the training efficiency.

\item  We conducted experiments on SNGCL for the node classification task, and the experimental results fully demonstrate its effectiveness. SNGCL outperforms the state-of-the-art unsupervised model in the node classification task.

\end{itemize}

%% 2.2
\section{Related Work}
\par{\noindent\textbf{\quad}}
\par{In this section, we briefly present recent advances in two related topics: contrast learning and unsupervised graph representation learning.}

\subsection{Contrast learning} 
\par{\noindent\textbf{\quad}}
\par{Contrast learning is a self-supervised learning technique that aims to learn feature representations by comparing samples with their similar and dissimilar counterparts to train the model. The technique is implemented by computing a contrast loss function, which motivates the model to bring the representations of similar samples closer together while pushing the representations of dissimilar samples farther apart. Compared to traditional supervised learning methods\textsuperscript{\cite{domingos2012few,breiman2001random,lecun1998gradient}}, contrast learning simplifies the model construction process, reduces the reliance on manually labeled data, and helps improve the model's capacity for generalization and its resilience. In the field of unsupervised representation learning, contrast learning has demonstrated excellent performance. Various contrast learning algorithms have been implemented. For example, the Deep Information Maximization (DIM)\textsuperscript{\cite{hjelm2018learning}} algorithm works to enhance the correlation between local and global representations of an image; the Contrastive Predictive Coding (CPC)\textsuperscript{\cite{oord2018representation}} algorithm predicts future outputs in the latent space through an autoregressive model that utilizes the current outputs as a priori to improve the feature representations; the Contrastive Multi-view Learning (CMC)\textsuperscript{\cite{tian2020contrastive}} algorithm aims to unsupervisedly learn the representations of a dataset from multiple perspectives; Simple Contrast Learning (SimCLR)\textsuperscript{\cite{chen2020simple}} algorithm improves feature representation by maximizing the mutual information between data-enhanced views and emphasizes the importance of constructing high-quality negative samples; Meanwhile, the SwAV\textsuperscript{\cite{caron2020unsupervised}} performs cluster-centered contrast learning based on clustering by integrating clustering algorithms and contrast learning concepts. The development of these algorithms has driven the application and advancement of contrast learning techniques in the field of self-supervised learning\textsuperscript{\cite{tang2024graphgpt}}. }
\begin{figure*}\label{F1}
	\centering
	  \includegraphics[width=\textwidth]{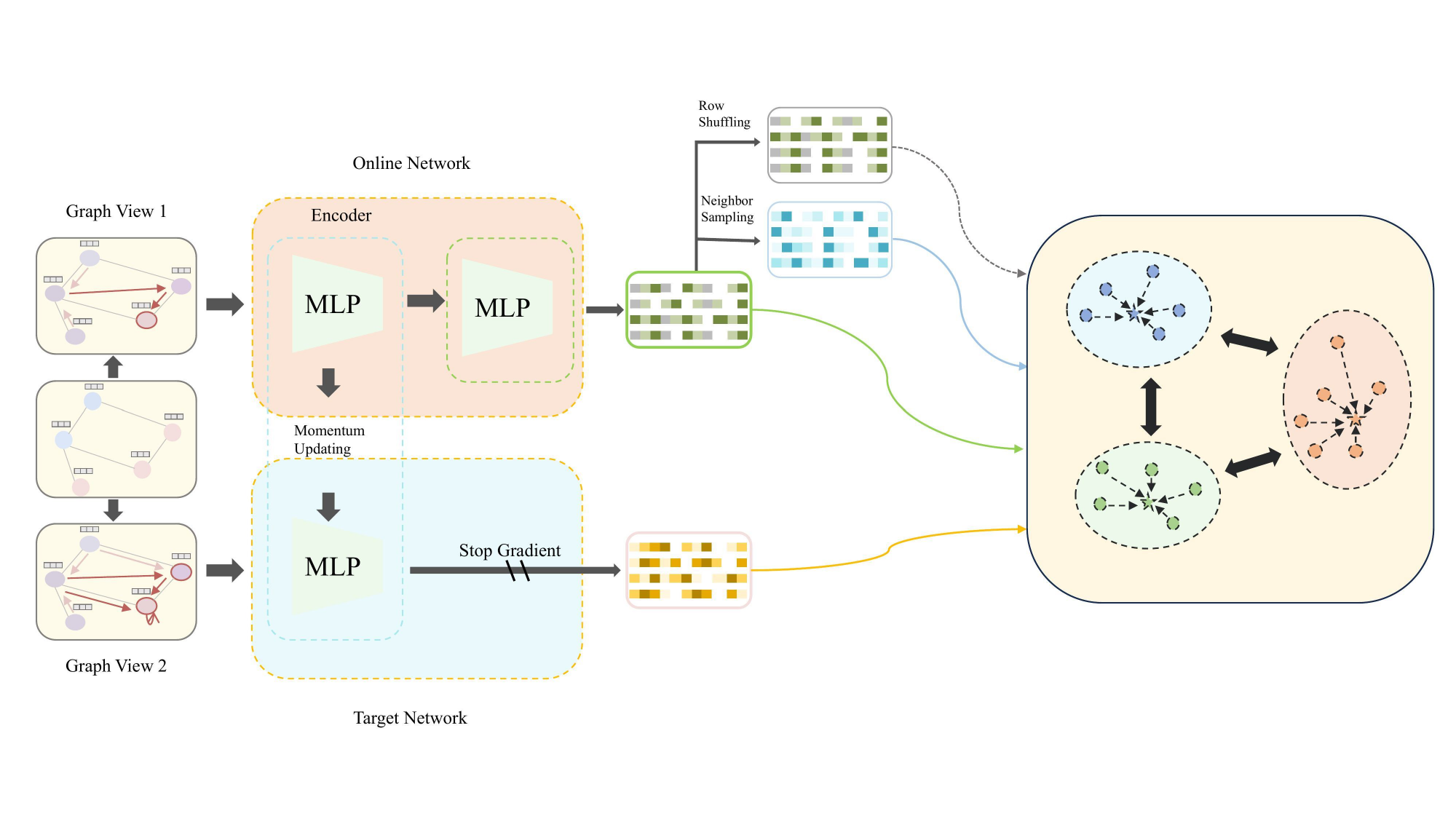}
	\caption{The framework of the proposed SNGCL model}
	\label{FIG:1}
\end{figure*}

%% 2.2
\subsection{Unsupervised graph representation learning}
\par{\noindent\textbf{\quad}}
\par{Traditional unsupervised graph representation learning techniques usually rely on random wandering or reconstruction of adjacency matrices, but these methods may encounter performance bottlenecks when dealing with large, high-dimensional and complex datasets, as well as performing discriminative tasks. To address these limitations, recently developed contrast learning techniques have shown greater potential to improve the strength and scalability of graph representation learning. Graph contrast learning, as a discriminative representation learning framework, is used to bring the representation of similar samples closer while making the representation of dissimilar samples more spread out by comparing the samples in the graph with similar and dissimilar samples and optimizing the model by comparing the loss functions.  }

\par{Some of these contrast learning methods include: DGI is a method for unsupervised learning of node representations within graphs, focusing on increasing the mutual information between enhanced graph representations and the graph data extracted. MVGRL trains graph encoders by maximizing the mutual information between graph-level representations encoded by different structural views.GMI\textsuperscript{\cite{peng2020graph}} distinguishes between positive- and negative-example graphs by maximizing the mutual information between graph-level representations of vectors and hidden representations. GRACE learns node representations by maximizing their consistency within attribute-level and structural-level views, improving the mutual information between input node features and high-level node embeddings.The core idea of BYOL\textsuperscript{\cite{grill2020bootstrap}} is to train an encoder network by constructing a comparative learning framework composed of two primary elements: a target network and an online network. Both networks are encoder networks whose goal is to map the input data to a low-dimensional representation in the latent space. BGRL\textsuperscript{\cite{thakoor2021bootstrapped}}, inspired by BYOL, introduces a self-supervised graphical representation that does not require negative samples, saving memory and computational costs. MERIT\textsuperscript{\cite{jin2021multi}}, inspired by siamese network, performs node-view comparisons through multiple scales. However these approaches face some challenges. First, as noted in surveys on graph data augmentation, many existing data augmentation methods can lead to significant differences between the generated view and the original view, affecting the view quality and model training effectiveness\textsuperscript{\cite{dong2024effectiveness,zhou2025data}}. At the same time, there are time-consuming and computationally resource-intensive problems in dealing with large-scale data. Second, most existing graph comparison learning methods significantly depend on large negative sample sets, which not only impose substantial computational overhead but may also, as recently argued, impair the model's capacity to learn semantically discriminative features\textsuperscript{\cite{huang2025does,saifuddin2025hypergcl}}.}

\section{Proposed Methods}
\par{\noindent\textbf{\quad}}
\par{Inspired by the application of self-supervised learning and its attribute graph clustering in comparison learning, we propose SNGCL a graph comparison learning model implementation for simple comparison node classification to solve the problem of data augmentation for generating new and old views in graph comparison learning that have large and time-consuming discrepancies and depend greatly on numerous negative samples.
Firstly, a Laplace smoothing filter with multiple layers is used as a step to process the data, and the filter is used as a low-pass filter to remove noise from the feature matrix's high-frequency components, and a normalized Laplace matrix and a symmetric normalized Laplace matrix are used to obtain the globally and locally smoothed feature matrices, respectively. The local and global feature matrices are passed into the backbone network, which employs a momentum-driven siamese network concatenation architecture to guide online e-learning to explore richer and better representations. Finally, two triple-configuration loss functions are employed to explore the complementary information between structural and neighborhood information, and two positive embeddings and their loss functions are generated by combining structural and neighborhood information to achieve larger inter-class differences. An upper bound is also used to ensure that the distance between the positive and anchor embeddings is finite to achieve smaller intra-class variation.

}
\par{The overall model framework is shown in \textbf{Figure 1}. It clearly shows the overall framework of our model. Next, we describe each step in detail.}

\subsection{Graph data}
\par{Given an attribute graph $G=(V, E, X)$, where $\mathcal{V}=\left\{v_{1}, v_{2}, \cdots, v_{n}\right\}$ is the set of vertices with $n$ nodes, $E$ is the set of edges, and $X=\left[\mathrm{x}_{1}, \mathrm{x}_{2}, \cdots, \mathrm{x}_{n}\right]^{T}$ is the feature matrix; the topology of the attribute graph $G=(V,E,X)$ is denoted by the adjacency matrix $\mathrm{A}=\left\{a_{i j}\right\} \in \mathbb{R}^{n \times n}$, and if $v_{i j}=1$, it indicates that there exists an edge between node $v_i$ to node $v_j$, and $D=\operatorname{diag}\left(d_{1}, d_{2}, \cdots, d_{n}\right) \in \mathbb{R}^{n \times n}$ denotes the degree matrix of $A$, where $d_{i}=\sum_{v_{j} \in \mathcal{V}} a_{i j}$ denotes the degree of node $v_i$.}

\subsection{Low-pass Denoising Operation}
\par{Given that the Laplace filter is able to achieve similar efficacy as the graph convolution operation, this paper cleverly utilizes this low-pass denoising operation for the aggregation of neighbor information. This step is designed as an independent preprocessing session before training, aiming at effectively filtering out the high-frequency noise in the attribute data, so as to enhance the stability and accuracy of the subsequent model training. In order to obtain a global view of feature smoothing, a normalized Laplace filter is used in this paper. The mathematical expression of this filter is:}

\begin{normalsize}
\begin{equation}\label{H1}
H^{r w}=I-L^{r w}
\end{equation}
\end{normalsize}

\par{where $I$ is the unit matrix and $L^{r w}$ denotes the normalized graph Laplace matrix, specifically $L^{r w}$ is expressed as:}

\begin{normalsize}
\begin{equation}\label{H2}
L^{r w}=\hat{D}^{-1} \hat{L}=I-\hat{D}^{-1} \hat{A}
\end{equation}
\end{normalsize}

\par{where $\hat{D}$ and $\hat{L}$ are the degree and Laplace matrices corresponding to $\hat{A}$, and $\hat{A}$ is $\hat{A}+I$. The localized eigen-smoothed view is obtained by a symmetric normalized Laplace filter with the filter equation:}

\begin{normalsize}
\begin{equation}\label{H3}
H^{sym}=I-L^{sym}
\end{equation}
\end{normalsize}

\par{where $L^{sym}$ represents the symmetric normalized Laplacian matrix of the graph, and the specific $L^{sym}$ representation is:}

\begin{normalsize}
\begin{equation}\label{H4}
L^{\text {sym }}=\hat{D}^{-1 / 2} \hat{L} \hat{D}^{-1 / 2}=I-\hat{D}^{-1 / 2} \hat{A} \hat{D}^{-1 / 2}
\end{equation}
\end{normalsize}

\par{Then, by superimposing t-layer normalized and symmetric normalized Laplace filters respectively, a filter consisting of multiple filters is obtained, and the purpose of this filter is to perform feature convolution operations on the graph data and extract richer and meaningful feature representations. By superimposing t-layer graph Laplace filters, the information from neighboring nodes is passed and aggregated layer by layer to obtain a more global and deeper feature representation.The superimposed t-layer normalized and symmetric normalized Laplace filters are publicized as follows:}

\begin{normalsize}
\begin{equation}\label{H5}
X_{s}^{r w}=\left(\prod_{i=1}^{t} H^{r w}\right) X=\left(H^{r w}\right)^{t} X
\end{equation}
\end{normalsize}

\begin{normalsize}
\begin{equation}\label{H6}
X_{s}^{sym}=\left(\prod_{i=1}^{t} H^{sym}\right) X=\left(H^{sym}\right)^{t} X
\end{equation}
\end{normalsize}

\par{where $X_{s}^{r w}$, $X_{s}^{sym}$ denote the smoothed feature matrix, and $\left(H^{r w}\right)^{t}$, $\left(H^{sym}\right)^{t}$ are stacked t-layer graph Laplace filters. This low-pass filtering operation removes high-frequency noise from attributes, enhancing training efficiency.}

\subsection{Structural Contrastive Module}
\par{In this section, using the momentum-driven siamese network architecture as the core framework, we take the smoothed global view and local view as the inputs to the target network and online network, respectively. To realize knowledge distillation, we apply the self-distillation method. }

\par{Firstly, in the online network, the smoothed feature matrix and adjacency matrix $\tilde{X}_{1}$, $\tilde{A}_{1}$, $Z^{1}=p_{\theta}\left(\tilde{X}_{1}, \tilde{A}_{1}\right)$ of the incoming local view are the coded representation of the MLP encoder that passes through the online network, and then the prediction of the online network, which is expressed as $H^{1}=q_{\theta}\left(Z^{1}\right)$. }

\par{In the target network, the smoothed feature matrices and adjacency matrices $\tilde{X}_{2}$, $\tilde{A}_{2}$, $Z^{2}=p_{\zeta}\left(\tilde{X}_{2}, \tilde{A}_{2}\right)$ of the incoming global view are the coded representations of the MLP encoder passing through the target network. In order to achieve the goal of bringing the representation of the same node from different views closer together, two networks are used in this paper to work together. This process not only distills knowledge from historical observations, but also ensures that the training process of the online network remains stable. The online network does not pass gradients directly to the target network during training. Instead, we apply a momentum update mechanism to update its parameters. The specific momentum update formula is as follows:}

\begin{normalsize}
\begin{equation}\label{H7}
\zeta^{t}=m \cdot \zeta^{t-1}+(1-m) \cdot \theta^{t}
\end{equation}
\end{normalsize}

\par{where $m$, $\zeta$ and $\theta$ are momentum, target network parameters and online network parameters, respectively.}

\subsection{ Comparative learning loss}
% \par{\noindent\textbf{\quad}}
\par{The output feature vector from the target network is denoted as $H^{+}$ (positive embedding), and the feature vector of the output feature vector from the online network is used as the anchor embedding, in which the feature vector generated by neighbor sampling of the anchor embedding is $\widetilde{H}^{+}$ (positive embedding), in which in order to obtain the positive embedding with the neighbor information, we firstly store the indexes of neighbor embeddings of all the nodes, and then sample them, and then compute the average of the samples. In this way, the node's neighboring information can be obtained efficiently with the following formula: }

\begin{normalsize}
\begin{equation}\label{H8}
\tilde{h}_{i}^{+}=\frac{1}{m} \sum_{j=1}^{m}\left\{h_{j} \mid v_{j} \in N_{i}\right\}
\end{equation}
\end{normalsize}

\par{Where $m$ is the number of sampled neighbors and $N_{i}$ is the set of 1-hop neighbors of node $v_{i}$. The feature vector generated by row shuffling of the anchor embedding is $H^{-}$ (negative embedding), which is publicized as follows:}

\begin{normalsize}
\begin{equation}\label{H9}
	H^{-}=\operatorname{Shuffle}\left(\left[h_{1}, h_{2}, \ldots, h_{N}\right]\right)
\end{equation}
\end{normalsize}

\par{The two positive embeddings generated consider the structural information and neighborhood information of the graph data, respectively.}

\par{In this paper, we explore the complementarity between structural and neighborhood information using two Triplet loss-based triple recombination loss functions, and an upper bound loss is designed to reduce the intra-class variation and expand the inter-class variation without the use of a discriminator. The triple recombination loss function for each sample can be expressed as: }

\begin{normalsize}
\begin{equation}\label{H10}
	\alpha+d\left(h, h^{+}\right)<d\left(h, h^{-}\right)
\end{equation}
\end{normalsize}

\par{where $d(,)$ is the similarity measure and $\alpha$ is a non-negative value to maintain a "safe" distance between positive and negative embeddings. For interclass analysis, summing over all negative samples, Eq. expands to:}

\begin{normalsize}
\begin{equation}\label{H11}
	L_{\text {triplet }}=\frac{1}{k} \sum_{i=1}^{k}\left\{d\left(h, h^{+}\right)^{2}-d\left(h, h_{i}^{-}\right)^{2}+\alpha\right\}_{+}
\end{equation}
\end{normalsize}

\par{Where $\{\cdot\}_{+}=\max \{\cdot, 0\}$, $k$ represents the number of negative samples.}

% \begin{figure}
% 	\centering
% 		\includegraphics[scale=.6]{./2.eps}
% 	\caption{Node-level Contrast}
% 	\label{F2}
% \end{figure}

\par{In order to increase the inter-class differences and widen the distance between positive and negative samples, the above operations are performed on the two positive embeddings $H^{+}$ and $\widetilde{H}^{+}$, denoted as $L_{S}$ and $L_{N}$, respectively:}

\begin{normalsize}
\begin{equation}\label{H12}
	L_{S}=\frac{1}{k}\sum_{i=1}^{k}\{d(h,h^{+})^{2}-d(h,h_{i}^{-})^{2}+\alpha\}_{+}
\end{equation}
\end{normalsize}

\begin{normalsize}
\begin{equation}\label{H13}
    L_{N}=\frac{1}{k}\sum_{j=1}^{k}\{d(h,\tilde{h}^{+})^{2}-d(h,h_{i}^{-})^{2}+\alpha\}_{+}
\end{equation}
\end{normalsize}

\par{For the intra-class analysis, an upper bound is used to ensure that the distance between the positive and anchor embeddings is finite, i.e., smaller intra-class differences are realized, with Eq:}
\begin{normalsize}
\begin{equation}\label{H14}
d\left(h, h^{-}\right)<d\left(h, h^{+}\right)+\alpha+\beta
\end{equation}
\end{normalsize}

\par{After summing the losses for all negative embeddings, the upper bound loss for reducing the within-class variation is given by Eq:}
\begin{normalsize}
\begin{equation}\label{H15}
L_{U}=-\frac{1}{k} \sum_{i=1}^{k}\left\{d\left(h, h^{+}\right)^{2}-d\left(h, h_{i}^{-}\right)^{2}+\alpha+\beta\right\}_{-}
\end{equation}
\end{normalsize}

\par{where $\{\cdot\}_{-}=\min \{\cdot, 0\}$. Finally the formula is integrated as:}
\begin{normalsize}
\begin{equation}\label{H16}
L=\omega_{1} L_{S}+\omega_{2} L_{N}+L_{U}
\end{equation}
\end{normalsize}

\par{where $\omega_{1}$ and $\omega_{2}$ are hyperparameters that regulate the degree of contribution of different losses to the final optimization objective, respectively.}

\renewcommand{\algorithmicrequire}{\textbf{Require:}} 
\renewcommand{\algorithmicensure}{\textbf{Output:}} 

\renewcommand{\thealgorithm}{\textbf{Algorithm 1:}}
    \begin{algorithm*}
        \caption{Procedure for training SNGCL}
        \begin{algorithmic}[1] \label{G1}
            \Require Graph characterized by node attribute matrix $\mathbf{X}$ and adjacency matrix $\mathbf{G}$, degree matrix $\mathbf{D}$, training epochs number $\mathbf{T}$, hyper-parameters $t$ ,learning rate and hyper-parameters $\omega_{1}$, $\omega_{2}$.
            \Ensure Node embedding $\mathrm{Z}$
            % \State xxxxx   
    \For {epoch =1 : T}
    \State Using a low-pass denoising module to obtain smoothed local and global feature matrices: $\tilde{X}_{1}$ and $\tilde{X}_{2}$ ;
    \State In Structural Contrastive Module, two embeddings are generated for the incoming $\left(\tilde{X}_{1}, \tilde{A}_{1}\right)$ and $\left(\tilde{X}_{2}, \tilde{A}_{2}\right)$ by the formula Eq.\eqref{H7}.;
    \State The embedding of the target network is used as the positive embedding $H^{+}$, the embedding of the online network is used as the anchor embedding $H$.;
    \State Generate negative embedding $H^{-}$ by row shuffling for anchor embedding by Eq.\eqref{H9} and positive embedding $\widetilde{H}^{+}$ by neighbor sampling for anchor embedding by Eq.\eqref{H8}.;
    
    \State To explore the structural and neighborhood information, two losses $L_{S}$ and $L_{N}$ are generated by Eq.\eqref{H11};
    
    \State Reducing within-class variation by using the ceiling loss designed by Eq.\eqref{H15};

    \State Optimize the entire framework according to the Eq.\eqref{H16};

   \EndFor
   % \Repeat
   % \State xxxx
   % \Until {xxx}
   % \State\Return xxx % Return 
\end{algorithmic}
\end{algorithm*}

\section{Experiments}

\subsection{Datasets} % 
\par{\noindent\textbf{\quad}}
\begin{table}
% \caption{Statistics of datasets.}
\footnotesize

\begin{tabular*}{\tblwidth}{@{} LLLLL@{} }

\textbf{Table 1}\\
Statistics of datasets.\\
\hline
\textbf{Dataset}      & \textbf{Nodes} & \textbf{Edges} & \textbf{Features} & \textbf{Communities} \\ \hline
\textbf{Cora}         & 2708           & 5429           & 1433              & 7                    \\
\textbf{Citeseer}     & 3327           & 4732           & 3703              & 6                    \\
\textbf{Pubmed}       & 19717          & 44338          & 500               & 3                    \\
\textbf{Amazon-Photo} & 7650           & 119081         & 745               & 8                    \\
\textbf{Coauthor-CS}  & 18333          & 81894          & 6805              & 15                   \\ \hline
\end{tabular*}
\label{T2}
\end{table}

\par{We carried out comprehensive experiments on five real-world datasets, including Cora, Citseeer, Pubmed, Amazon Photo, and Coauthor CS. \textbf{Table 1} briefly summarizes the experimental datasets.}

\par{Cora: The Cora dataset is a dataset of 2708 scientific publications with 5429 edges connecting them. This dataset has 7 categories (artificial intelligence, machine learning, etc.). Each publication is represented by a word vector consisting of 0s and 1s. The full dictionary contains 1433 distinct words. }

\par{CiteSeer: The CiteSeer dataset includes 3312 scientific publications that are categorized into six different categories. 4732 links exist in this dataset, which constitute the citation network. In total, 3703 unique words make up the dictionary.}

\par{PubMed: The PubMed dataset contains 19,717 scientific publications from the Pubmed database of studies in the field of diabetes. These publications were categorized into three categories, namely medicine, biology and chemistry. And they form a citation network between them, which contains 44338 links. Each publication is identified by a keyword represented by a 500-dimensional word vector.}

\par{Amazon Photo: The Amazon Photo dataset is part of the Amazon dataset, which contains 7,650 nodes and 119,081 edges. These edges represent a situation where two products are frequently bought together. All goods can be categorized into 8 different categories. Each good is identified by a keyword represented by 745 feature vectors.}

\par{Coauthor CS: The Coauthor CS dataset is a part of the Coauthor dataset, which contains 18,333 authors and 81,894 edges. These edges represent common collaborative relationships between authors. All authors can be categorized into 15 different research areas. Each author consists of 6,805 features.}

\subsection{Evaluation metrics} 
\par{\noindent\textbf{\quad}}
\par{In order to measure the effectiveness and accuracy of the model proposed in this paper, it needs to be tested and evaluated in an effective way. In this paper, ACC (Accuracy.), a commonly used evaluation metric for node classification, is selected.}
\par{ACC compares the predicted labels with the real labels provided by the data and is defined as shown in Eq.\eqref{H16}.}

\begin{normalsize}
\begin{equation}\label{H17}
	A C C=\frac{\sum_{i=1}^{N} \delta\left(y_{i}, \operatorname{map}\left(r_{i}\right)\right)}{N}
\end{equation}
\end{normalsize}

\par{Where $r_{i}$ is the classified label and $y_{i}$ is the true label. $n$ is the total number of data and the map function denotes the reproduction assignment of the best class label to ensure correct statistics. $\delta$ denotes the indicator function as follows:}

\begin{normalsize}
\begin{equation}\label{H18}
	\delta(t, z)=\left\{\begin{array}{l}
1, t=z \\
0, t \neq z
\end{array}\right.
\end{equation}
\end{normalsize}

\subsection{Baselines} 
\par{\noindent\textbf{\quad}}
\par{We compared the SNGCL to ten competing methods. The details are as follows:}
\par{GCN\citep{kipf2016semi}: GCN generates new node representations by combining node information through edge data. By performing a convolution operation between the features of a node and its neighboring nodes, the node's representation is updated at each layer, and each node's representation is influenced by its neighboring nodes.}

\par{GAT\citep{velickovic2017graph}: GAT automatically learns the impact of neighboring nodes on the central node using the attention mechanism to determine the weights. All the neighbors of each node are given an attention coefficient which are normalized to represent the importance of the neighboring nodes on the central node. A new representation of the center node is obtained by weighted summation.}
\par{DGI\citep{velickovic2019deep}: DGI uses an existing GCN as a graph encoder to learn the representation of a node by maximizing the mutual information (MI) between the neighboring representations and the corresponding higher-order graph summaries. It enables node representations to preserve the structural information of neighboring nodes as much as possible through self-supervised learning.} 

\par{GMI\citep{peng2020graph}: GMI is a method to assess the link between the input graph and its advanced hidden representation, which measures the common information in terms of both node features and topology.GMI is able to preserve and extract rich information about the graph structure data into the embedding space.}
\par{GIC\citep{mavromatis2020graph}: GIC uses k-means clustering on top of node representations to get cluster level aggregation. It maximizes the mutual information of both the graph set aggregation and the cluster level aggregation to learn the node representation.}
\par{GRACE\citep{zhu2020deep}: GRACE learns node representations by randomly generating two views on the original graph and using a node-level contrast loss function to bring closer the same node embedding in both views.} 

\begin{table*}[width=\textwidth]
\renewcommand{\arraystretch}{1.1}
  \begin{tabular*}{\tblwidth}{@{} LLLLLLLLLL@{} }
\textbf{Table 2}\\
 Experimental performance of node classification \\
\hline
Model       & Training Data & Cora     & Citeseer & Pubmed   & Amazon-Photo & Coauthor-CS \\ \hline
GCN         & X,A,Y         & 81.5     & 70.3     & 79       & 85.8         & 92.4        \\
GAT         & X,A,Y         & 83.7$\pm$0.7 & 72.5$\pm$0.7 & 79.0$\pm$0.3 & 84.8$\pm$0.5     & 90.9$\pm$0.1    \\
DGI         & X,A           & 81.8$\pm$0.2 & 71.5$\pm$0.2 & 77.3$\pm$0.6 & 80.5$\pm$0.2     & 91.0$\pm$0.0    \\
GMI         & X,A           & 82.7$\pm$0.2 & 73.0$\pm$0.3 & 80.1$\pm$0.2 & 83.9$\pm$0.0     & 91.7$\pm$0.0    \\
GIC         & X,A           & 83.1$\pm$0.3 & 72.3$\pm$0.5 & 80.2$\pm$0.3 & 89.4$\pm$0.1     & 91.8$\pm$0.0    \\
GRACE       & X,A           & 80.0$\pm$0.4 & 71.7$\pm$0.6 & 79.5$\pm$1.1 & 81.8$\pm$1.0     & 90.1$\pm$0.8    \\
MVGRL       & X,A           & 82.9$\pm$0.7 & 73.7$\pm$0.1 & 80.8$\pm$0.1 & 85.8$\pm$0.5     & 89.6$\pm$0.1    \\
MERIT       & X,A           & 83.5$\pm$0.2 & \textbf{74.0$\pm$0.3} & 80.8$\pm$0.2 & 87.5$\pm$0.3     & 92.6$\pm$0.1    \\
SUGRL(2022) & X,A           & 83.4$\pm$0.5 & 73.0$\pm$0.4 & 81.9$\pm$0.3 & 87.3$\pm$0.1     & \textbf{93.2$\pm$0.4}    \\
NCLA(2023) & X,A           & 82.2$\pm$1.6 & 71.7$\pm$0.9 & \textbf{82.0$\pm$1.4} & 90.2$\pm$1.3     & 91.5$\pm$0.7    \\

SNGCL      & X,A           & \textbf{84.0$\pm$0.2} & 72.9$\pm$0.3 & 80.8$\pm$0.1 & \textbf{93.2$\pm$0.4}     & 92.4$\pm$0.3    \\ \hline
\end{tabular*}
\end{table*}

\par{MVGRL\citep{hassani2020contrastive}: MVGRL learns node and graph representations by using a diffusion technique on the original graph and sampling to get two views, then maximizing the mutual information between one view's node representation and another view's graph representation.}
\par{MERIT\citep{jin2021multi}: MERIT generates two augmented views from the original graph, based on local and global views. It then employs a cross-view and cross-network contrastive objective to maximize the consistency between node representations from different views and networks.}
\par{SUGRL\citep{mo2022simple}: SUGRL devises a method to remove GCN-generated anchors and negative embeddings, and uses a triple-group loss function to explore complementary information between structural and neighbor information to amplify inter-class variation. In addition, it uses ceiling loss to reduce intra-class variation.} 

\par{NCLA\citep{shen2023neighbor}: NCLA introduces an innovative method that integrates graph data augmentation with contrastive learning to enhance node representations through the study of advanced graph structures. It concentrates on the neighboring information of nodes and utilizes contrastive learning to strengthen the similarities among nodes.}

\subsection{Node Classification Performance} 
\par{\noindent\textbf{\quad}}

\par{In this section, we conduct experiments on node classification to evaluate the performance on these five experimental datasets. Among the three citation networks, we randomly selected 20 nodes for each class as training data and used the training data for learning to find the optimal parameters. Then 500 nodes are randomly selected as validation data, and the rest are used as test data. For the other two datasets, we randomly select 30 nodes for each class as training data and validation data respectively. And the remaining nodes are taken as test data.}
\par{The results are presented in \textbf{Table 2}., where we use bold to indicate the optimal values. SNGCL outperforms the baseline method on five datasets. It outperforms the suboptimal model by 1\% on the Cora dataset. Especially, excellent results are obtained on the Amazon-Photo dataset, while it improves 5.8\%, 6\% compared to MERIT and SUGRL, respectively, and 3.1\% improvement compared to the latest NCLA. This validates the effectiveness of the SNGCL algorithm. From the experiment outcomes, the following conclusions can be drawn:}
\par{From the \textbf{Table 2}, it can be seen that in most cases in the real network dataset, the algorithm results of SNGCL are better than other baseline algorithms. It has better performance. It is worth observing that MERIT has a better performance on the Citeseer dataset, and SUGRL has a better performance on the Coauthor-CS dataset, but SNGCL still has improvement compared to the other algorithms, and has a large advantage over traditional GCN. Observation of other datasets reveals that on the Cora dataset, SNGCL improves the algorithm performance by 1\% compared to SUGRL. On the Pubmed dataset, however, SNGCL also achieves better results than the baseline algorithm. Especially on the Amazon-Photo dataset excellent results are obtained, then compared to MERIT, SUGRL improved by 5.8\%, 6\% respectively. Based on the above analysis, it can be concluded that from the experimental results, the SNGCL algorithm performs well on the real dataset and is competitive with other algorithms.}
\par{To balance the trade-off between bias and variance and enhance the model's generalization capability, a model ensemble strategy was implemented during the fine-tuning phase. This approach ensures that SNGCL maintains robust predictive performance across diverse graph topographies, further contributing to the superior accuracy observed in the comparative study.}

\begin{figure}
	\centering
	  \includegraphics[width=0.53\textwidth]{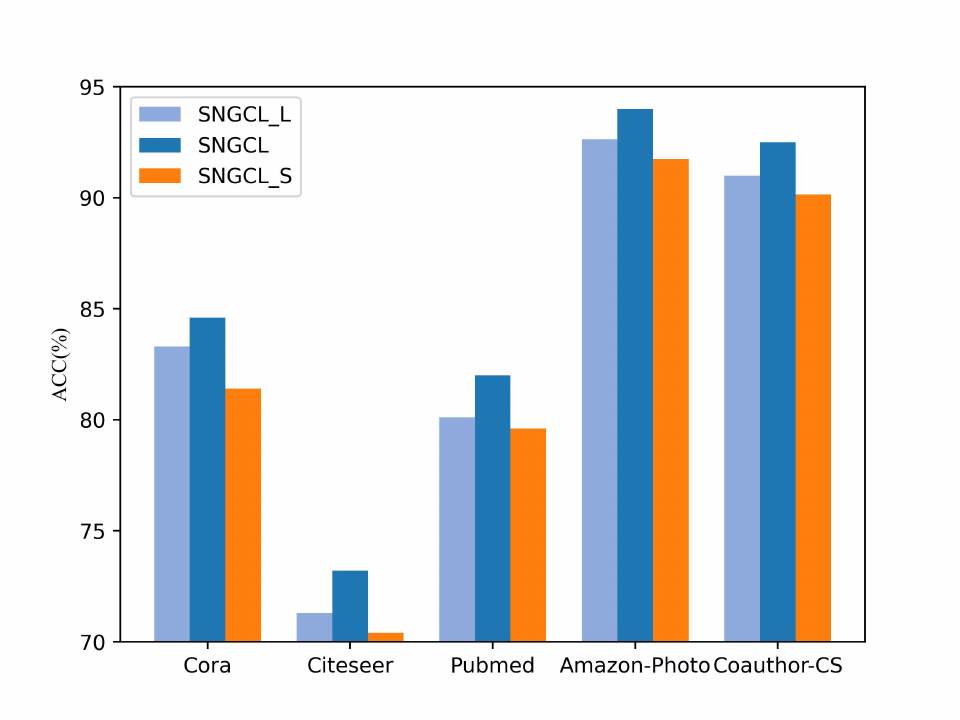}
	\caption{The results of ablation study.}
	\label{F6}
\end{figure}

\begin{enumerate}[(1)]
\item Compared to DGI, GIC outperforms DGI in some aspects by exploring the cluster-level node information in the graph and utilizing the global information more efficiently.However, the performance improvement of GIC is not particularly impressive when compared to other models. As an example, MERIT, MVGRL, and SUGRL all show strong performance on the node classification task.MERIT shows a more significant advantage over MVGRL by maximizing the alignment between global and local node embeddings, which emphasizes the importance of consistency in node representation. Although MVGRL relies only on the underlying global information, its results are still competitive, which proves the critical role of global information.SUGRL employs clustering to shrink the distance between nodes of the same class and expand the distance between nodes of different classes, which highlights the equal importance of strengthening intra-network connectivity and relaxing inter-network connectivity.
\item SNGCL outperforms SUGRL. demonstrates the effectiveness of using superimposed t-layer normalized Laplace matrix and symmetric normalized Laplace matrix to obtain globally and locally smoothed feature matrices, respectively. Compared with SUGRL, SNGCL generates higher quality graph representations and has better performance in downstream task node classification.

\end{enumerate}

\subsection{Ablation study} 
\par{\noindent\textbf{\quad}}

\par{We have done ablation experiments on SNGCL and also verified the effectiveness of each module. In this paper, the algorithm in the low-pass denoising module uses stacked t-layer normalized Laplace matrices and symmetric normalized Laplace matrices, respectively, to obtain global and locally smoothed feature matrices, which are passed into the siamese network, in order to improve the accuracy of the algorithm. Following the taxonomy of graph data augmentation \textsuperscript{\cite{Ding2022DataAugmentation}}, we dissect our augmentation module into structure-oriented (global) and feature-oriented (local) components. Therefore, in order to verify the accuracy of the proposed algorithm more comprehensively, experiments are conducted using the global-only feature matrix algorithm SNGCL\_L and the local-only feature matrix algorithm SNGCL\_S, respectively, to analyze their effects on the algorithm. The accuracy rate (ACC) is adopted as the evaluation index, and the experiments are conducted on five datasets respectively.}
\begin{figure}
	\centering 
	\includegraphics[width=0.53\textwidth]{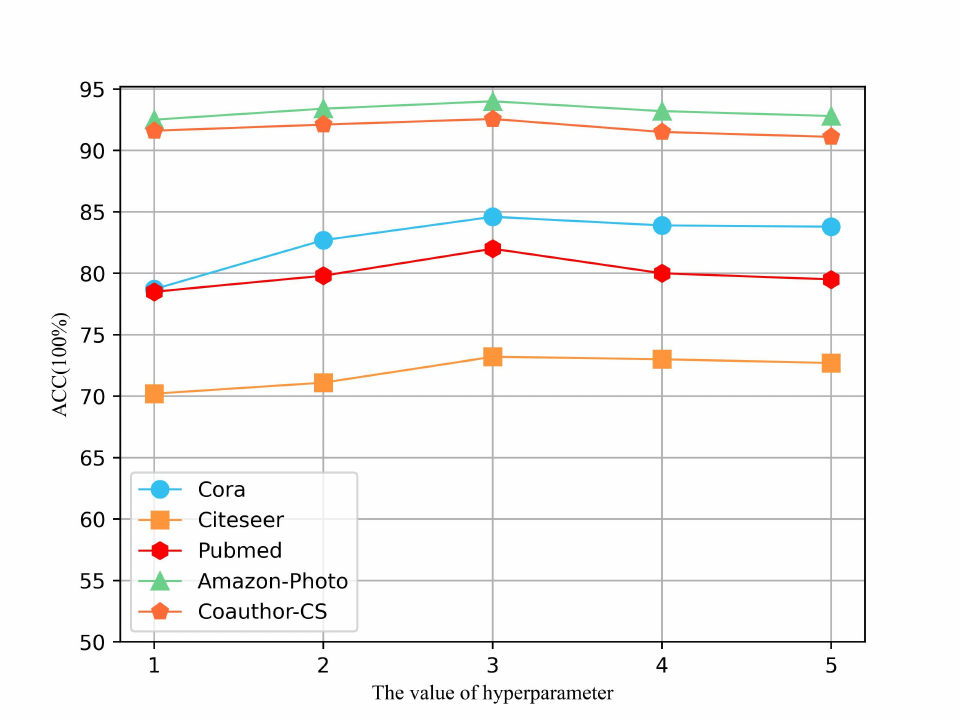}
	\caption{Analysis of the stacking layer parameter $t$.}
	\label{F5}

\end{figure}

\begin{figure*}
	\centering
	  \includegraphics[width=\textwidth]{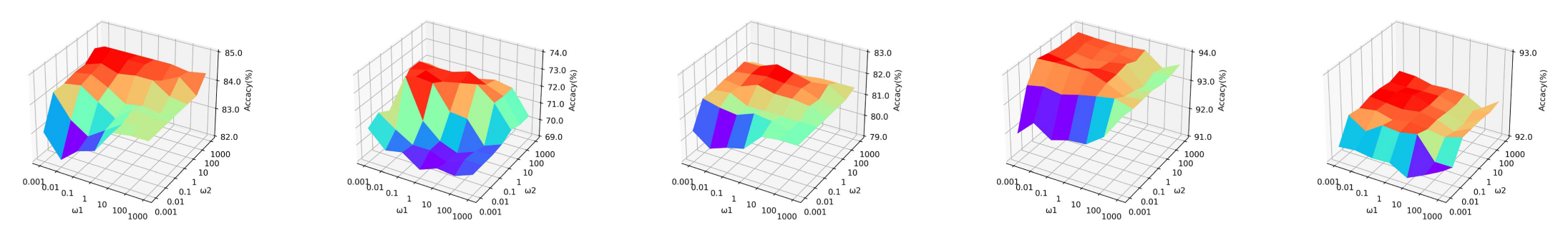}
	\caption{Experimental result graph of hyperparameter $\omega_{1}$ and $\omega_{2}$
}
	\label{F7}

\end{figure*}
\par{The results are shown in \textbf{Figure 2}, and we can find that the performance of our model decreases without this key component on the five datasets, which proves the effectiveness of our global and local smoothing feature matrices passed into the siamese network. Specifically, our proposed model can improve SNGCL\_L by 1.9\%, 1.4\% and SNGCL\_S by 1.8\% and 1.2\% on the CiteSeer and Amazon-Photo datasets, respectively.}

\subsection{ Parameters study} 
\par{\noindent\textbf{\quad}}
\par{\noindent\textbf{ Stacking layer parameter}}
\par{In the data enhancement module, this paper investigates the effect of Laplace filter with superimposed t layers on the experiments, where the number of superimposed layers t is taken from 1 to 5 and the ablation experiments are performed on 5 datasets. The final experimental results are shown in \textbf{Figure 3}, using the accuracy (ACC) as the evaluation metric. In the figure, it can be observed that the number of superimposed layers t achieves the optimal result in all five datasets in terms of accuracy when the value is taken as 3, and then the accuracy starts to decrease gently. When too many layers are superimposed, the Laplace filter keeps smoothing the image. This may result in details and edge information in the image being over-smoothed out, making the image blurry and losing its original features. There is also a risk of reduced model stability, meaning that the model may produce large output changes for small changes in the input data.}

\par{\noindent\textbf{\quad}}
\par{\noindent\textbf{ Compare loss parameters}}
\par{In this paper, we investigate the effect of hyperparameters in SNGCL, i.e., $\omega_{1}$ and $\omega_{2}$ in Eq.\eqref{H15}. node classification is performed by varying the values of $\omega_{1}$ and $\omega_{2}$ from $10_{-3}$ to $10_{3}$ on five datasets, setting the momentum update parameter m to 0.8 while fixing the weight of LU to 1, and then reporting the results as shown in \textbf{Figure 4}. If the values of $\omega_{1}$ and $\omega_{2}$ are too small ($10_{-3}$ and $10_{-2}$), the accuracy is reduced. This suggests that both LS and LN are important as they help to further push the positive embedding away from the negative embedding.}

\subsection{Visualization} 
\par{\noindent\textbf{\quad}}

%\par{To show the superiority of our models, we visualize the results of the DeepWalk, DGI and SNGCL models performing the node classification task on the Cora and Citeseer datasets by means of the t-SNE\citep{van2008visualizing} algorithm, where the node colors denote different categories. In \textbf{Figure 5}, the 2D projections of SNGCL present a clearer separation, suggesting that our approach facilitates the graph encoder to extract more expressive node representations for the node classification task.}

\par{To demonstrate the superiority of the proposed model, we employ the $t$-SNE algorithm \citep{van2008visualizing} to visualize the node representations learned by DeepWalk, GCN, and SNGCL on the Cora and Citeseer datasets. As illustrated in \textbf{Figure 5}, where node colors signify distinct categories, the first and second rows correspond to the experimental results on Cora and Citeseer, respectively. It can be observed that the 2D projections of SNGCL exhibit clearer class separation and more compact clusters compared to the baselines. This observation reinforces that our SNGCL framework effectively facilitates the graph encoder in extracting more expressive and discriminative node representations, which are highly conducive to the node classification task.}

\begin{figure*}
	\centering
	  \includegraphics[width=0.8\textwidth]{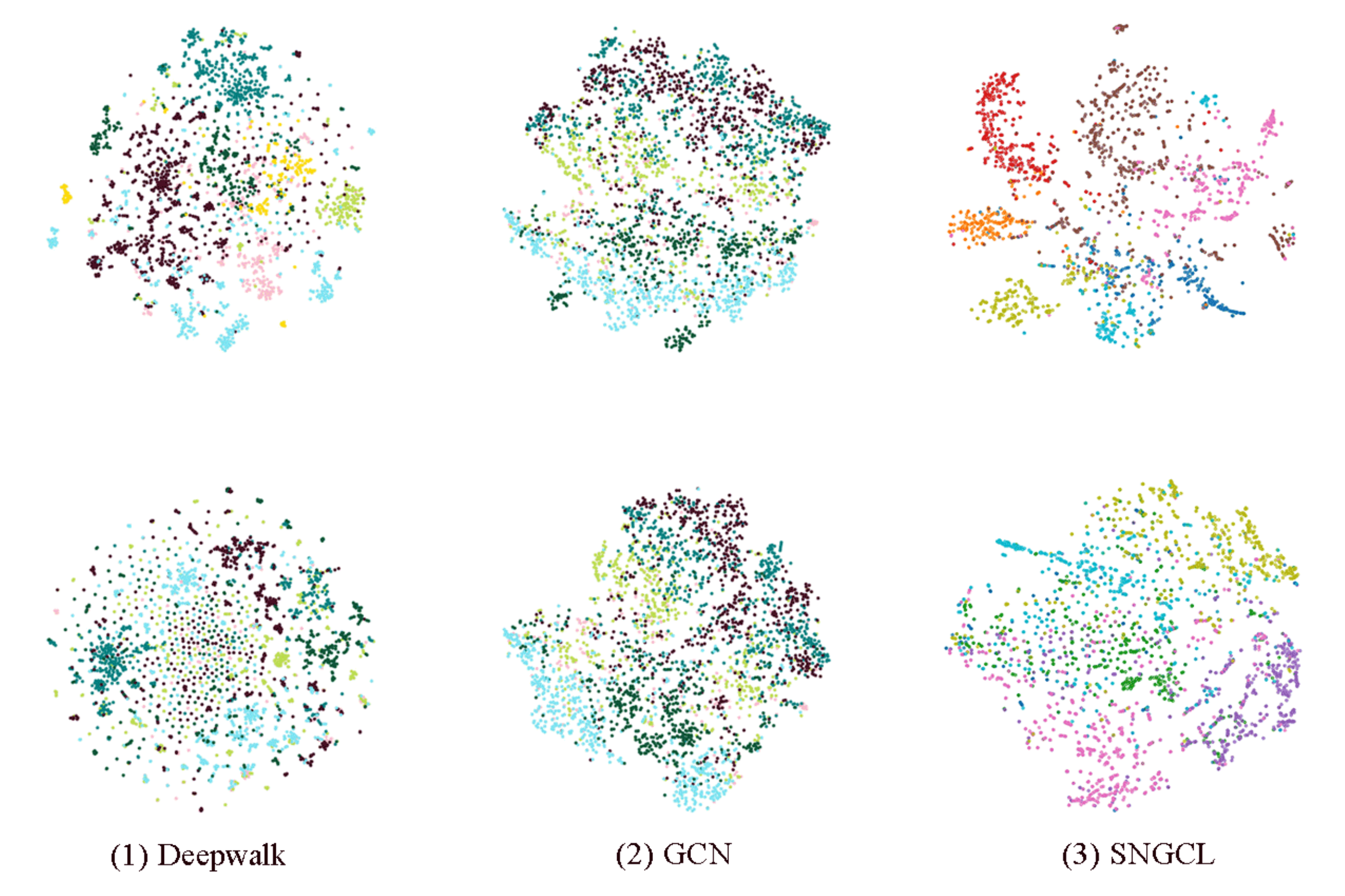}
	\caption{Visualization of experimental results.
}
	\label{F7}

\end{figure*}

\section{Conclusion}
\par{\noindent\textbf{\quad}}

\par{In this paper, we propose a SNGCL where we propose the use of a Laplace smoothing filter as a step in processing the data, obtaining the global and local feature smoothing matrices, respectively, which are thus passed into the target and online networks of the siamese network, and finally employing a modified triple recombination loss function to close the intraclass distances and to distance the interclass distances. Experimental results show that SNGCL outperforms state-of-the-art methods on several datasets for the node classification task, proving the superiority and effectiveness of the method.}

% \section*{Acknowledgments}

% This work was jointly supported by the following projects: (1)the National Social Science Fund of China under Grant 18BGL266 and Grant 20BSH076. (2)the National Nature Science Foundation of China (41971365).

% Here are two sample references: \citep{Fortunato2010}
% \citep{Fortunato2010,NewmanGirvan2004}
% \citep{Fortunato2010,Vehlowetal2013}

% \section{Floats}
% {Figures} may be included using the command,\linebreak 
% \verb+\includegraphics+ in
% combination with or without its several options to further control
% graphic. \verb+\includegraphics+ is provided by {graphic[s,x].sty}
% which is part of any standard \LaTeX{} distribution.
% {graphicx.sty} is loaded by default. \LaTeX{} accepts figures in
% the postscript format while pdf\LaTeX{} accepts {*.pdf},
% {*.mps} (metapost), {*.jpg} and {*.png} formats. 
% pdf\LaTeX{} does not accept graphic files in the postscript format. 


% \bibliographystyle{cas-model2-names}
\bibliographystyle{unsrt}

\bibliography{cas-refs}


% Author biography without author photo.
% Author biography. Author biography. Author biography.
% Author biography. Author biography. Author biography.
% Author biography. Author biography. Author biography.
% Author biography. Author biography. Author biography.
% Author biography. Author biography. Author biography.
% Author biography. Author biography. Author biography.
% Author biography. Author biography. Author biography.
% Author biography. Author biography. Author biography.
% Author biography. Author biography. Author biography.

% \bio{figs/pic1}
% Author biography with author photo.
% Author biography. Author biography. Author biography.
% Author biography. Author biography. Author biography.
% Author biography. Author biography. Author biography.
% Author biography. Author biography. Author biography.
% Author biography. Author biography. Author biography.
% Author biography. Author biography. Author biography.
% Author biography. Author biography. Author biography.
% Author biography. Author biography. Author biography.
% Author biography. Author biography. Author biography.

% \bio{figs/pic1}
% Author biography with author photo.
% Author biography. Author biography. Author biography.
% Author biography. Author biography. Author biography.
% Author biography. Author biography. Author biography.
% Author biography. Author biography. Author biography.

\end{document}